%% file: iclr2025_conference.tex
\documentclass{article} % For LaTeX2e
\usepackage{iclr2025_conference,times}
\usepackage{amsmath}
\usepackage{algorithm}
\usepackage{algpseudocode}
\usepackage{pifont}
\usepackage{graphicx} 
\usepackage{xcolor}
\usepackage{tcolorbox}
\usepackage{booktabs}
\usepackage{threeparttable}
\usepackage{svg}
\usepackage{array}
\usepackage{subcaption}
\usepackage{multirow}
\usepackage{booktabs} % 用于更漂亮的表格线条

\input{math_commands.tex}

\usepackage{hyperref}
\usepackage{url}

\title{OpenRFT: Adapting Reasoning Foundation Model for Domain-specific Tasks with Reinforcement Fine-Tuning}

\author{Yuxiang Zhang, Yuqi Yang, Jiangming Shu, Yuhang Wang, Jinlin Xiao \& Jitao Sang\thanks{Corresponding author.} \\
School of Computer Science and Technology\\
Beijing Jiaotong University\\
Beijing, China\\ 
\\
\texttt{jtsang@bjtu.edu.cn} \\
\small
}

\iclrfinalcopy % Uncomment for camera-ready version, but NOT for submission.
\begin{document}

\maketitle
\begin{abstract}
OpenAI's recent introduction of Reinforcement Fine-Tuning (RFT) showcases the potential of reasoning foundation model and offers a new paradigm for fine-tuning beyond simple pattern imitation. This technical report presents \emph{OpenRFT}, our attempt to fine-tune generalist reasoning models for domain-specific tasks under the same settings as RFT. OpenRFT addresses two key challenges of lacking reasoning step data and the limited quantity of training samples, by leveraging the domain-specific samples in three ways: question augmentation, synthesizing reasoning-process data, and few-shot ICL. The evaluation is conducted on SciKnowEval, where OpenRFT achieves notable performance gains with only $100$ domain-specific samples for each task. More experimental results will be updated continuously in later versions. Source codes, datasets, and models are disclosed at: \href{https://github.com/ADaM-BJTU/OpenRFT}{https://github.com/ADaM-BJTU/OpenRFT}.

\end{abstract}

\section{Introduction}
OpenAI's o1 model has shown strong reasoning abilities in mathematics and programming, but its generalization to other tasks remains uncertain. The recent introduction of Reinforcement Fine-Tuning (RFT)~\citep{openai_rft} has provided a promising avenue for reasoning generalization. With only dozens of high-quality $(question, answer)$ pairs, RFT enables the creation of customized reasoning models excelling at domain-specific tasks. 

The significance of RFT is at least two-fold: (1) It demonstrates the promise of using generalist reasoning models, like o1, as reasoning foundation models. By enabling the efficient creation of domain-specific reasoning models, RFT practically expands the applicability of reasoning models across diverse tasks.  (2) It introduces a new paradigm for fine-tuning foundation models. Unlike Supervised Fine-Tuning (SFT), which merely mimics patterns in training data, RFT leverages reasoning capabilities to facilitate thinking and trial-and-error learning. This brings models closer to achieving human-like generalization, moving beyond mechanical imitation to extrapolate knowledge to new cases.

It is believed that the core techniques behind RFT are closely related to those of o1. Inspired by recent o1-replication efforts~\citep{openo1,openr_2024,llama_o1_2024,zhao2024marcoo1openreasoningmodels,zhang2024o1codero1replicationcoding}, we attempt to develop an implementation under the same settings as the RFT demo, which we call \emph{OpenRFT}. While this early exploration may not achieve optimal results, we hope it is beneficial to the community for clarifying the conceptual landscape and inspiring further advancements in this area. 

%data augmentation: rephrasing, synthesize, 回放
%prm enhanced RL
%domain knowledge embedding

Realizing RFT requires addressing two key challenges: the absence of reasoning step data in the provided domain-specific samples, and the limited quantity of such samples. For the first challenge, lacking reasoning process supervision may lead to rollout data where the final outcome is correct, but the reasoning steps are flawed (example illustrated in Fig.~\ref{fig:badcase} in \emph{Appendix}). This will introduce incorrect reward signals, causing an imbalance between exploration and exploitation. OpenRFT addresses this challenge with two approaches. ~(1) Reasoning process synthesis and SFT: We begin by prompting the vanila model to roll out and fill in the missing reasoning steps in the domain-specific samples. These synthesized data are then used to fine-tune the original reasoning foundation model via SFT. This allows the policy model to adapt to the reasoning process of the domain task, providing a more robust starting point for the subsequent RL phase. ~(2) Incorporating a Process Reward Model (PRM) in RL: PRM helps supervise the rationality of the reasoning process, enhance the probability of correct reasoning process rollouts, and thus stabilize the RL training.

For the second challenge, Reinforcement Learning (RL) generates data by exploring the environment and adapting its learning distribution. This reduces reliance on the initial sample quantity. However, having only dozens or hundreds of samples may still be insufficient. OpenRFT addresses this with two approaches. ~(1) Data augmentation: This approach directly increases the data volume by rephrasing questions and shuffling options to generate new domain-specific samples. ~(2) Domain knowledge embedding: This approach aims to enhance the efficiency of RL exploration. We introduce a simple prompting-based technique: utilizing domain-specific samples in a few-shot In-Context Learning (ICL) setup to guide the policy model's exploration.

We selected reasoning foundation model and process reward model from the Skywork-o1 series~\citep{skyworkopeno12024}. The evaluation is conducted on SciKnowEval~\citep{feng2024sciknoweval}, a newly-released scientific benchmark that encompasses five distinct abilities. We have chosen $8$ specific tasks corresponding to the reasoning ability at level L3, spanning various disciplines including biology, chemistry, physics, and materials science. Experimental results show that using only $100$ domain-specific samples, OpenRFT increases performance by an average of $11\%$. We also find that more data, a stronger reasoning foundation model, and better alignment of the action space all contribute to improved results. 

It is important to note that, much like 
how an effective System-2 model (e.g., o1) relies on a powerful System-1 model (e.g., GPT-4o), the feasibility of RFT depends on having a strong generalist reasoning model and a corresponding PRM. Only with such models can the reasoning process for domain-specific tasks be understood and scored. Since there is currently no open-source, highly effective generalist reasoning model, we merely offer an early implementation of RFT. It is believed that when stronger general reasoning models emerge, the full potential of RFT will be further unlocked.
%

%Furthermore, we have adapted the PRM with domain knowledge to enhance its suitability for specific tasks.

%Additionally, we also adapted the PRM using domain knowledge to further align it with the specific reasoning requirements of the target tasks. This adaptation strengthens the integration of domain-specific insights into the reasoning process.

%%In reinforcement fine-tuning, you provide the system the results that it is supposed to achieve. The system will figure out the reasoning process needed to achieve the objective.

%我们提出引入domain-specific knowledge来预先adapt PRM（和policy model），提高RL效率。
%adapt方法:prompt 中加入domain-specific context
%-role setting
%-CoT schema 
%-few-shot ICL（+knn）
%-RL通过逐步改变生成的数据分布（也是自己要学习的），从而影响策略模型

%%与正式模型训练完全一样，相当于沿着基础模型的训练方式在领域数据上继续训练
%%可选：1、是否先生成CoT和对policy model进行SFT；2、prm是否同步更新，进行selfplay

%注：adapt PRM时可以用很长prompt，因为解码很短
%adapt policy model时

\section{Method}
\begin{figure}[t] % 插入图片的位置，htbp分别代表这里、顶部、底部和页面底部
  \centering % 使图片居中
  \includegraphics[width=0.98\textwidth]{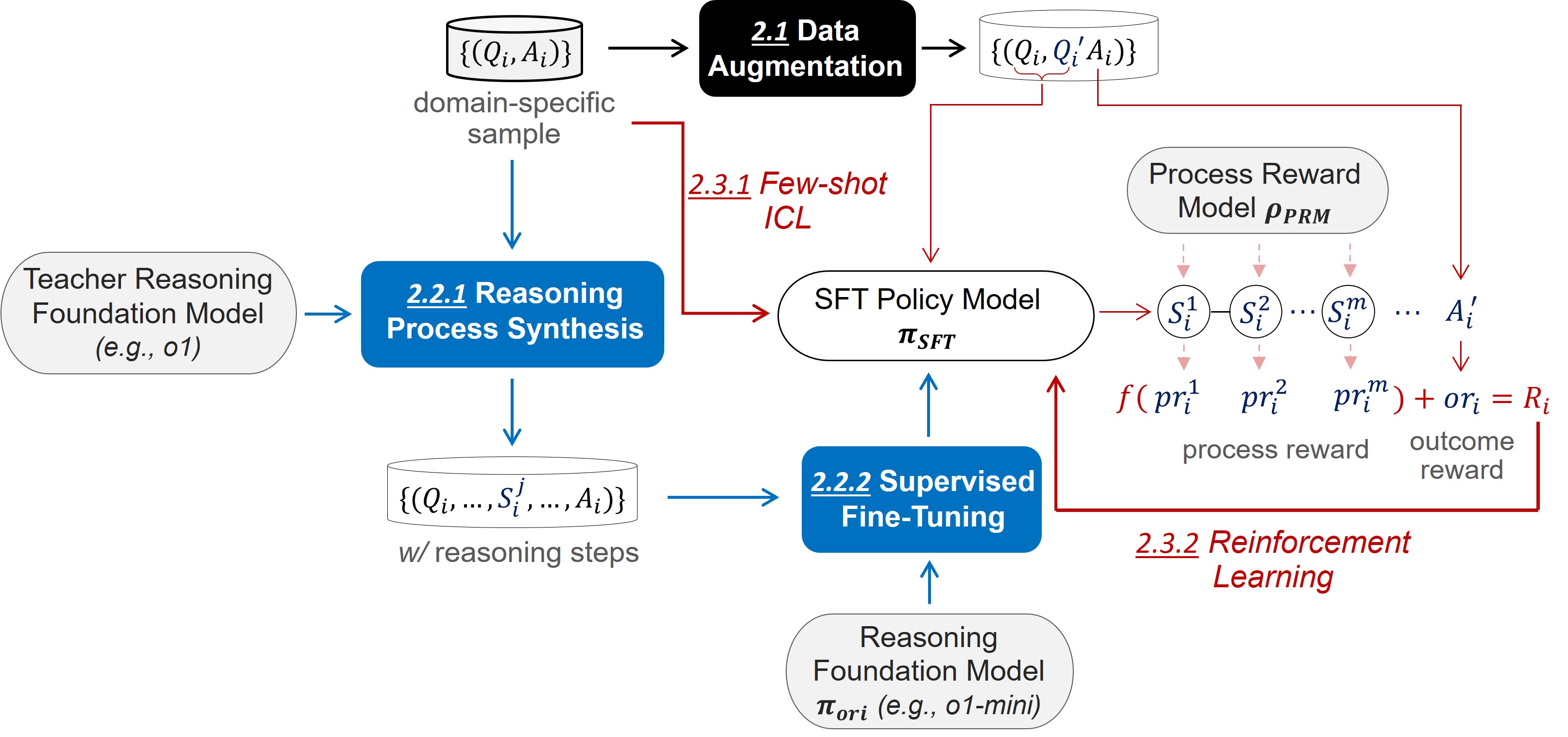} % 插入图片，设置图片宽度为文本宽度的一半
  \caption{OpenRFT framework.} % 图片的标题
  \label{fig:2} % 图片的标签，用于文中引用
\end{figure}

The key to Reinforcement Fine-Tuning lies in the effective utilization of limited domain-specific samples. Based on the way domain-specific samples are leveraged, as illustrated in Fig.~\ref{fig:2}, the framework of OpenRFT can be divided into three modules. ~(1) \emph{Data augmentation}: By rewriting questions and shuffling options, we explicitly generate more domain-specific data. This helps explore a broader range of states and actions in the RL stage. ~(2) \emph{SFT-based imitation}: Using a stronger reasoning foundation model as a teacher~\footnote{Typically, a smaller reasoning foundation model (e.g., o1-mini) is desired to ensure efficiency in domain-specific applications. When synthesizing reasoning step data, it is ideal to use a stronger reasoning foundation model as the teacher model for distillation (e.g., o1). It is important to ensure that the action space of the teacher and student models remains consistent.

Due to the lack of a stronger reasoning model with consistent actions, in our reported experiments, the synthesis is instead performed by the policy model itself.}, the missing reasoning steps are synthesized for the provided domain-specific data. These enhanced samples are then used to pre-adapt the student policy model through SFT. ~(3) \emph{RL-based exploration and self-improvement}: The domain-specific samples are incorporated into the policy model in a few-shot ICL manner. The policy model, under process supervision by the PRM, explores and continuously optimizes within an RL environment.

\subsection{Data Augmentation}

Data augmentation (DA) is a widely used technique for addressing data scarcity. In this framework, we propose a data augmentation method based on question rewriting. Specifically, we first utilize GPT-4o-mini to rewrite the question stem while preserving its original meaning, generating multiple variations. Subsequently, the options for each question are shuffled independently for these variations. Both the original question and the newly transformed question are used simultaneously in the PPO optimization.

\begin{figure}[H]
\centering
\begin{tcolorbox}[colframe=black!75!white, colback=white!95!black, boxrule=0.75mm, arc=5mm, outer arc=5mm, title=Task Instructions for Data Augmentation]

\textbf{Task:} \\
Given a paragraph, generate five distinct expressions while preserving the original meaning.

\bigskip

\textbf{Instructions:}
\begin{itemize}
    \item The paragraph is a scientific multiple-choice question (without options).
    \item Keep the meaning unchanged; do not add extra or unrelated information.
    \item Adjust sentence structure if necessary, but the meaning must remain the same.
    \item Provide five variations, each separated by "\textless sep \textgreater".
\end{itemize}

\bigskip

\textbf{Question:} \\
\texttt{\{question\_paragraph\}}

\end{tcolorbox}

\caption{Task instructions for generating distinct expressions}
\label{fig:task_instructions}
\end{figure}

\subsection{SFT-based Imitation}

\subsubsection{Reasoning Process Synthesis} \label{sec:data-synthesis}
To enhance the understanding and response capabilities of a reasoning foundation model in specific domains, we adopt a teacher reasoning foundation model to synthesize reasoning process data and leverage this data for SFT of the reasoning foundational model. The teacher model is typically a stronger general reasoning foundation model capable of synthesizing high-quality reasoning samples tailored to domain-specific data. 

Specifically, for each question \(Q_i\) in the domain dataset\((Q_i,A_i)\), We designed a set of general prompts to leverage the reasoning capabilities of the teacher model, enabling it to generate initial reasoning processes and answers within its action space. To ensure the correctness and diversity of the generated data, we introduce a multi-sampling strategy to synthesize high-confidence reasoning process data \((Q_i, \ldots, S_i^j, \ldots, A_i^{\prime})\) , where \(S_i^j\) represents the \(j\)-th sampled intermediate reasoning step generated by the teacher model. By sampling multiple reasoning paths for each question \(Q_i\), we select at least one data that can infer the correct answer \(A_i\).

For each question \(Q_i\), the final synthesized reasoning data \((Q_i, S_i, A_i)\) includes the question, the intermediate reasoning steps \(S_i\), and the corresponding answer \(A_i^{\prime}\). These data points are then aggregated to form the domain-specific reasoning dataset \(\mathcal{D}_{\text{process}}\)  , which serves as a high-quality resource for SFT of the reasoning foundation model. 

\subsubsection{SFT of Policy Model}

After completing the reasoning process synthesis tasks described in Section~\ref{sec:data-synthesis}, we use each complete reasoning solution in the dataset to initialize the policy model \(\pi_{ori}\). This step aims to let the policy model \(\pi_{ori}\) learn and generalize reasoning patterns and strategies embedded in \(\mathcal{D}_{\text{process}}\) , thereby enabling it to generate accurate and coherent reasoning solutions for new, unseen questions within the domain.

Given the question \(Q_i\), the policy model \(\pi_{ori}\) utilizes the reasoning dataset \(\mathcal{D}_{\text{process}}\) to predict intermediate reasoning steps and derive a final answer. Specifically, \(\pi_{ori}\) is trained to maximize the likelihood of producing the synthesized reasoning paths and correct answers from \(\mathcal{D}_{\text{process}}\). The training objective is formulated as:

\[
\mathcal{L}_{\text{SFT}} = -\sum\nolimits_{(Q_i, S_i, A_i) \in \mathcal{D}_{\text{process}}} \log P(S_i, A_i | Q_i; \pi_{ori}),
\]

where \(P(S_i, A_i | Q_i; \pi_{ori})\) represents the probability of the policy model generating the reasoning steps \(S_i\) and answer \(A_i\) conditioned on the question \(Q_i\).

During training, the model learns to balance the trade-off between mimicking the high-quality reasoning paths synthesized by the teacher model and generalizing to new questions. 

After the SFT process, we obtain the updated policy model \(\pi_{SFT}\). This model incorporates the reasoning patterns, strategies, and domain-specific knowledge embedded in the dataset \(\mathcal{D}_{\text{process}}\).

\subsection{RL-based Exploration and Self-improvement}
\subsubsection{Few-shot ICL-based Domain Knowledge Embedding}
In scenarios where there are only a few dozen training samples within a specific domain, the policy model lacking domain knowledge may face inefficiencies during the exploration stage. To address this, we employ a prompt-based in-context learning paradigm, with the aim of supplementing the policy model with knowledge and steering it toward reasoning and answering in the correct direction.

For each training task, we construct a corresponding vector database. For each training sample \( Q_i \), we retrieve the top \( k \) most similar question-answer pairs \(\{(Q_j, A_j)\ |\ j=1, 2, \dots, k\}\) based on vector similarity. These pairs are then concatenated with the task instructions to create an enhanced prompt enriched with domain-specific knowledge.

\subsubsection{RL with PRM}
The reward feedback based only on the outcome is easily influenced by the correct answer sampled in reinforcement learning, but with incorrect processes. This phenomenon is more likely to occur in long-range reasoning tasks. To mitigate the negative impact of this phenomenon to some extent, we integrate a process-based reward model $\rho_{\text{PRM}}$ into the reward design of reinforcement learning.

In the reinforcement learning phase, the policy model $\pi_{\theta}$ improves itself using domain-specific datasets constructed under different data organization schemes. We model the entire sampling as a language-augmented Markov Decision Process(MDP), formally represented as $\mathcal{M}=(\mathcal{V}, \mathcal{S}, \mathcal{A}, \mathcal{T}, \mathcal{R})$\citep{wang2024openr, carta2023grounding}. $\mathcal{V}$ represents the vocabulary, while the action space $\mathcal{A} \subseteq \mathcal{V}^N$ and state space $\mathcal{S} \subseteq \mathcal{V}^N$ are both made up of sequences of tokens. In this framework, $s_0$ represents the given question $Q_i$, and the action $a_t$ is considered as either a reasoning step or the final answer prediction (referring to $S_i^t$ as mentioned in Figure \ref{fig:2}). Here, different actions $a_t$ are separated by newline characters. The state transition function $\mathcal{T}: \mathcal{S} \times \mathcal{A} \rightarrow \mathcal{S}$ defines how the current state $s_t \in \mathcal{S}$ changes when an action $a_t \in \mathcal{A}$ is taken. Specifically, it concatenates the tokens in action $a_t$ directly to the current state $s_t$ to form a new state $s_{t+1} = \mathcal{T}(s_t, a_t)$. The reward function $\mathcal{R}: \mathcal{S} \times \mathcal{A} \rightarrow \mathbb{R}^+$ consists of both outcome-based rewards and process-based rewards.

For the final action in the sampling process, where the policy model $\pi_{\theta}$ produces the answer prediction $A_i'$, we compare it with the standard answer $(Q_i, A_i)$ from the domain dataset to obtain the outcome reward score $or_i$. If $A_i'$ matches $A_i$, the score is set to 1, otherwise it is set to 0. For all actions $a_t$ except for the final answer prediction, we use the process reward model $\rho_{\text{PRM}}$ to generate the process reward score $pr_i^t = \rho_{\text{PRM}}(s_{t-1}, a_t)$ for each reasoning step. We then combine the process rewards and the outcome reward $or_i$ to get the final reward score $R_i$ for the sample $(Q_i, A_i)$ using the recomputing function $f(\cdot)$ and a linear combination:
$$
R_i = \alpha \times or_i + (1 - \alpha) \times f(pr_i^1, pr_i^2, \cdots, pr_i^m)
$$
Here, the function $f(\cdot)$ represents different reward aggregation strategies, such as mean, minimum, etc. We use the Proximal Policy Optimization (PPO)\citep{schulman2017proximal} algorithm to train the policy model $\pi_{\theta}$ based on this reward function definition. The optimization objective is:
$$
\max_{\theta} \mathbb{E} \left[ \min \left( r(\theta) \hat{A}, \text{clip}(r(\theta), 1 - \epsilon, 1 + \epsilon) \hat{A} \right) \right]
$$
where $r$ calculates the probability ratio between the new and old policy. The advantage estimate $\hat{A}$ is crucial for guiding updates and is computed as the difference between the expected future return under the current policy and the baseline or value function. This advantage is directly influenced by the reward $R$, which makes the optimization process closely related to both the accuracy and rationality of the answer prediction and the reasoning process.

\section{Experiments}

\subsection{Datasets}
The experimental data in our study is sourced from the knowledge reasoning level~(L3) of the dataset provided in SciKnowEval~\cite{feng2024sciknoweval}, a newly-released scientific benchmark that encompasses five distinct abilities.
In this work, we focus on $8$ subsets, named \textit{GB1-ftness-prediction, retrosynthesis, chemical-calculation, molecule-structure-prediction, high-school-physics-calculation, material-calculation, diffusion-rate-analysis, perovskite-stability-prediction} (denoted as \emph{T1}-\emph{T8}), covering four domains: \textit{Biology, Chemistry, Physics, and Materials}. 
The questions in these subsets are in the form of multiple-choice questions (e.g., ABCD).
To simulate the practical scenario of fine-tuning with limited training samples, we sample 100 training samples and 100 testing samples from each dataset~\footnote{Since \textit{diffusion-rate-analysis} contains only 149 examples, the number of testing samples for this task is 49.}.

\subsection{Comparison Methods}
% \textcolor{red}{Vanilla, SFT, or None?}
The following methods are introduced as our comparative approaches. In addition to these, we also report the results of o1-mini and gpt-4o-mini, which can be regarded as a strong ceiling performance.

\begin{itemize}
    \item \textbf{Vanilla.} The policy model, i.e.,  Skywork-o1-Open-Llama-3.1-8B, which is used without any modifications or adjustments.
    \item \textbf{SFT.} The experimental setup involves distilling process data with limited training samples using the policy model, and then performing SFT on the policy model to obtain the final model.
    \item \textbf{SFT+.} 
    SFT with distillation data from a stronger model, QwQ-32B-Preview~\cite{qwq-32b-preview, qwen2}.
    % Note that since this model does not always produce reasoning process that matches the correct answer through sampling, if the correct-answer reasoning process is not obtained after more than 64 sampling attempts, we only provide the final answer without including the reasoning process.
    \item \textbf{ReFT.}  The method is derived from \cite{luong2024reft}, but without the warm-up stage. The model is obtained by directly training the vanilla policy model and the outcome reward model through PPO reinforcement learning on a limited set of training samples.
    \item \textbf{ReFT + PRM.} The method is built upon the ReFT, with the addition of a process reward model to enhance the reward signal.
    \item \textbf{SFT + RL(PRM).} The policy model is first initialized using SFT with the data from self-distillation, followed by reinforcement learning with process rewerd. 
    \item \textbf{SFT + RL(PRM) + DA.} This method builds upon the \emph{SFT + RL(PRM)} by incorporating data augmentation during reinforcement learning, thereby expanding the scale of the reinforcement training dataset.

    % \item \textbf{SFT + RL(PRM) + DA.} The method first applies data augmentation to the limited training samples, then uses self-distilled data for supervised fine-tuning, and finally performs process supervision through reinforcement learning.
    % \item \textbf{SFT + RL(PRM) + ICL} 
    % This approach aims to enhance the efficiency of RL exploration by utilizing domain-specific samples in a few-shot In-Context Learning (ICL) setup to guide the policy model’s exploration.
    \item \textbf{SFT+RL(PRM)+DA+ICL.} This implements all the proposed modules. First, we use data augmentation techniques to increase the number of available training samples. Then, we initialize the policy model by performing SFT on the process data during self-distillation.
    Finally, we finetune the model using process supervision reinforcement learning, and employ ICL techniques to guide RL exploration.
    
    % \item \textbf{OpenRFT (SFT+RL(PRM)+DA+ICL)}
    % \item \textbf{OpenRFT+(policy model enhanced)}
\end{itemize}

\subsection{Experimental Setup}
We conduct experiments using the foundation models from the Skywork o1 Open series~\cite{skyworkopeno12024}. 
These models demonstrate strong cognitive reasoning abilities in tasks involving mathematics, code, and other reasoning tasks.
Specifically, we employ Skywork-o1-Open-Llama-3.1-8B as the policy model and Skywork-o1-Open-PRM-Qwen-2.5-7B as the process reward model~\footnote{In proprietary implementations, such as OpenAI, the policy model and PRM are expected to be aligned, sharing the same action space and state-transition. However, in typical experimental settings, it is more common that only an open-source generalist reasoning model is available acting as the policy model, while the corresponding PRM is often inaccessible. 

In our current implementation, both the policy model and the PRM are from the Skywork-o1 series~\citep{skyworkopeno12024}.  However, the model provider has not explicitly stated that their action spaces are aligned, which may potentially impact the performance of RFT.}.

\noindent\textbf{Hyper-parameters}\par
% 内容大纲 参考 ReFT
In all experiments, training is conducted on devices equipped with NVIDIA H20-96GB GPUs, utilizing OpenRLHF~\cite{hu2024openrlhf} for reinforcement learning. Both SFT and RL experiments employ LoRA fine-tuning with a rank of 4.

For SFT training, the batch size, learning rate, number of epochs, and maximum sequence length are set to 8, 5e-5, 8, and 2048, respectively. In the PPO setup, the actor and critic learning rates are set to 3e-5 and 6e-5, respectively. Additionally, the maximum generated sequence length is set to 1536, and the KL coefficient is fixed at 0.01. The weighted coefficient, $\alpha$, for combining the process and outcome reward models is set to 0.7.

Regarding data augmentation, the original dataset is expanded by a factor of six, yielding a total of 600 samples. For synthetic reasoning process data, roll-out sampling is performed on each training sample until a response that correctly matches the true answer is obtained. This reasoning data is then treated as the ground truth for the corresponding training sample. In cases where 64 sampling attempts do not yield the correct answer, particularly when using the QwQ-32B-Preview model, the true answer is directly used as the response in the training data, bypassing the reasoning process data. This strategy is also applied in the Skywork-o1 experiments.

For in-context learning (ICL), Sentence-BERT is used to calculate the similarity between the current question and existing examples. The top 3 most similar examples are then selected as context for the current query.

Finally, for model generation, the temperature coefficient is set to 0.6 for all models, except for o1-mini, where the official guidelines specify a temperature coefficient of 1. All other generation parameters are set to their default values.

\noindent\textbf{Evaluation}\par
The answer output format is explicitly defined in the prompt to facilitate the extraction of answers from the model's output using predefined rules. This approach allows us to directly compare the model’s predictions with the ground truth.
We set the maximum sampling length to 2,048. For samples where the answer cannot be identified within this length, we consider the prediction as incorrect in the calculation.
We report accuracy values for both methods across all datasets.
Since the GPT-4o-mini and o1-mini models are sufficiently robust, we report the results based on a single evaluation. For other methods, we perform three evaluations and report the average accuracy.

% \subsection{Settings}
%/////dataset
% SciKnowEval~\citep{feng2024sciknoweval}
% -task information
% -training/testing set details

% \emph{GB1 ftness prediction}, \emph{retro synthesis}, \emph{chemical calculation}, \emph{molecule structure prediction}, \emph{high school physics calculation},
% \emph{material calculation}, 
% \emph{diffusion rate analysis},
% \emph{perovskite stability prediction} (denoted as \emph{T1}-\emph{T8}).

%/////implementation details
% Skywork-o1 series~\citep{skyworkopeno12024}.

%when introducing PRM, add:

%/////baselines
% GPT-4o, o1-mini, ReFT (only RL, w/o PRM \& SFT)

\begin{table}[t]
\centering
    \small
    \resizebox{\textwidth}{!}{
    \begin{tabular}{c c |c c c| c |ccc|c }
    \toprule[1.5pt]
    \multirow{2}{*}{\textbf{Model/ Method}} & \multicolumn{1}{c}{\textbf{Biology}} & \multicolumn{3}{c}{\textbf{Chemistry}}& \multicolumn{1}{c}{\textbf{Physics}} & \multicolumn{3}{c}{\textbf{Materials}} & \multirow{2}{*}{\textbf{\emph{Avg.}}}  \\ 
    \cmidrule(lr){2-2} \cmidrule(lr){3-5} \cmidrule(lr){6-6} \cmidrule(lr){7-9}
     &  T1 & T2 & T3 & T4 & T5 & T6 & T7 & T8 & \\ 
    \midrule
    % GPT-4o  & \textbf{0.50} & \textbf{0.86} & 0.82 & \textbf{0.36} & 0.64 & 0.58 & 0.79 & \textbf{0.57} & \textbf{0.641}\\ 
    GPT-4o-mini & \textbf{0.37} & 0.69 & 0.84 & \textbf{0.32} & 0.53 & 0.49 & 0.90 & 0.525 & 0.583 \\
    o1-mini & 0.35 & \textbf{0.86} & \textbf{0.87} & 0.23 & \textbf{0.73} & \textbf{0.70} & \textbf{0.87} & 0.50 & \textbf{0.639}\\ 
    \midrule
    Vanilla & 0.28 & 0.55 & 0.52 & 0.23 & 0.45 & 0.34 & 0.41 & 0.41 & 0.403 \\
    ReFT & 0.27 & 0.50 & 0.52 & 0.23  & 0.44 &0.33 & 0.41 &0.50 & 0.402 \\
    ReFT+PRM & 0.30 & 0.57 & 0.49 & 0.23 & 0.44 & 0.36 & 0.37 & 0.48 & 0.405 \\\midrule
    SFT & \underline{0.33} & 0.53 & 0.49 & 0.20 & 0.45 & 0.37 & 0.43 & 0.49 & 0.415\\
    SFT+RL(PRM) & 0.29  & 0.59 & 0.52 & 0.24 &\underline{0.47} & 0.36 & 0.46 & 0.57 & 0.437 \\
    SFT+RL(PRM)+DA & 0.29 & 0.63 & \underline{0.53} & 0.21 & \underline{0.47} & \underline{0.38} & 0.48 & \underline{\textbf{0.59}} & \underline{0.447}\\

        {SFT+RL(PRM)+DA+ICL} & \underline{0.33} & 0.57 & 0.52 & \underline{0.28} & 0.46 & 0.36 & \underline{0.49} & 0.53 & {0.443} \\
        
    \bottomrule[1.5pt]
    \end{tabular}
    }
  \caption{Accuracy of different models/methods. \textbf{Bold} indicates the highest value, while \underline{underline} indicates the highest value among the different methods based on the open-source Skywork-o1.} %添加此处
  \label{table:1}
\end{table}

\subsection{Results}
The main results are summarized in Table~\ref{table:1}.
% ~(1) o1-mini demonstrated the strongest reasoning capabilities, yet \emph{GPT-4o} showed a competitive performance in certain tasks. As the representative of System-1 models, \emph{GPT-4o} excels in versatility, outperforming o1-mini on tasks where domain knowledge is crucial. 
% ~(1) o1-mini and \emph{GPT-4o} exhibit strong and comparable performance across almost all tasks. \emph{GPT-4o}, as a strong System-1 model, outperforms o1-mini in tasks that require domain-specific knowledge. This indicates that both knowledge and reasoning are crucial in knowledge-based reasoning tasks. 
Key observations include: ~(1) o1-mini demonstrated the strongest reasoning capabilities, yet GPT-4o-mini showed a competitive performance in certain tasks. As the representative of System-1 models, GPT-4o-mini excels in versatility, outperforming o1-mini on tasks where domain knowledge is crucial. ~(2) The contribution of \emph{ReFT} is trivial. Designed to enhance general reasoning abilities, it fails to address the distribution discrepancy between the provided domain samples and the policy model to be fine-tuned. ~(3) With PRM to supervise the reasoning process, \emph{ReFT+PRM} contributes to  increasing the likelihood of sampling correct reasoning processes, although the improvement is not significant. ~(4) After fine-tuning with self-synthesized reasoning process data, \emph{SFT} slightly outperformed \emph{Vanilla}. This indicates that with such a limited number of samples, relying solely on SFT is far from sufficient. However, it can provide a good exploration starting point for subsequent RL. ~(5) Compared with \emph{SFT}, \emph{SFT+RL(PRM)} shows obvious improvement, highlighting the necessity of RL in fully leveraging the limited domain-specific samples. ~(6) \emph{SFT+RL(PRM)+DA} achieves consistent improvement over \emph{SFT+RL(PRM)} in different tasks, validating the contribution of domain-specific samples in synthesizing new samples. It achieves the best performance among the methods initialized with Skywork-o1, an average improvement of 11\% compared to \emph{Vanilla}. ~(7) By further incorporating few-shot ICL, there was no improvement but a slight decrease in the performance. This is possibly due to the inconsistent prompts during the SFT and RL stages. It is interesting to see that few-shot ICL benefits the most challenging task (T4: \emph{molecule-structure-prediction}), which is precisely where GPT-4o-mini excels. This somewhat underscores the effectiveness of domain knowledge. We are experimenting alternative ways for domain knowledge embedding.

% GPT-4o vs. o1-mini: 通用vs推理
% ReFT not good
% ReFT vs. ReFT+PRM: stablizing
% SFT vs. vanilla: warm-up, but not good enough with limited data (need RL to explore)
% OpenRFT: demonstrates the effectiveness of..
% OpenRFT-: SFT+RL(PRM) ~ ablation study (w/o data augmentation and few-shot ICL)

\subsection{Discussions}
%/////数据数量

% ReFT, SFT, OpenRFT
\begin{figure}[t] % 插入图片的位置，htbp分别代表这里、顶部、底部和页面底部
  \centering % 使图片居中
  \includegraphics[width=0.6\textwidth]{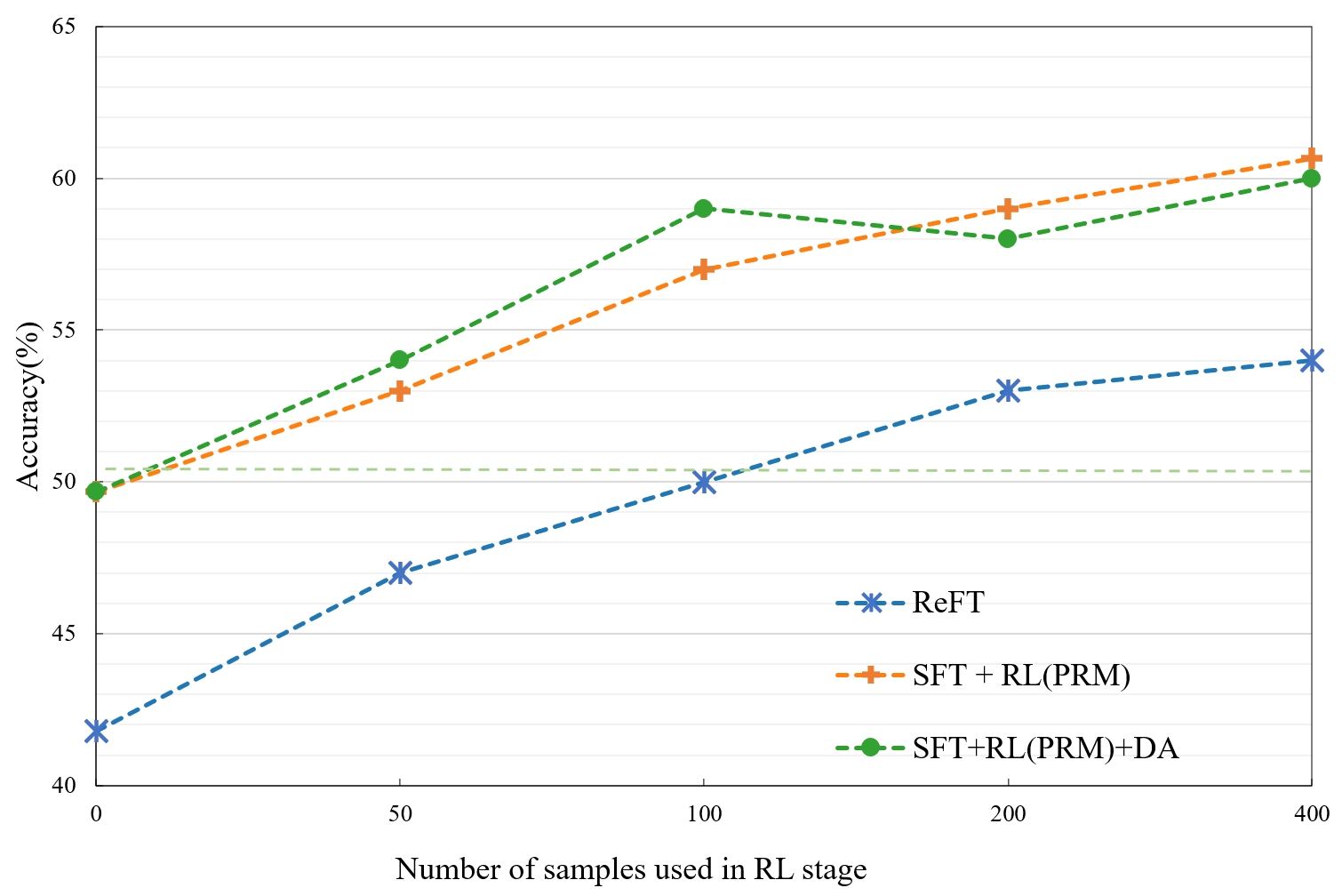} % 插入图片，设置图片宽度为文本宽度的一半
  \caption{Performance with different sizes of domain-specific data. The light green dashed line represents the performance of SFT with 100 samples.} % 图片的标题
  \label{fig:data_scale} % 图片的标签，用于文中引用
\end{figure}

\hspace{3mm} \textbf{More domain-specific data contributes to better RFT results.}
% To evaluate the impact of data size on the ReFT method, we compared the effects of varying data sizes on ReFT and two variants of OpenRFT using the T8 dataset. In all experiments involving OpenRFT variants, the models were initialized with an identical set of 100 samples for all conditions.
From Table~\ref{table:1}, we observe that data augmentation techniques significantly enhance model performance when training data is limited. 
This suggests that more domain-specific datasets could further improve the \emph{ReFT} method. 
Motivated by this, we explored the impact of data size on \emph{ReFT} by comparing the effects of different data sizes on \emph{ReFT} and two variants of \emph{OpenRFT} using the T8 dataset. 
In all experiments with the \emph{OpenRFT} variants, the models are initialized with the same set of 100 samples across all conditions.

As shown in Fig. \ref{fig:data_scale}, the performance of all three methods improves as the number of training samples increases. The influence of data augmentation is most pronounced when the training dataset is small. However, as the size of the data set increases, the benefit of data augmentation diminishes. This reduction in effectiveness may be due to the inherent errors introduced by LLM-based data augmentation, particularly when handling complex molecular formulas or similarly challenging scenarios.

%/////teacher policy vs. student policy alignment
\vspace{2mm}
\hspace{3mm} \textbf{Teacher policy model should align with the student policy model.} ~As introduced in Section~2, it is ideal to use a stronger reasoning foundation model as the teacher model to synthesize the reasoning process for the domain-specific samples. We employed the more powerful QwQ-32B as the teacher model, and then performed SFT on the student policy model using the synthesized reasoning process data. 

Table~\ref{table:2} displays the performance of the student policy model when generating its own process data (referred to as \emph{SFT}) and when using the process data synthesized by QwQ-32B-Preview~\citep{qwq-32b-preview} (referred to as \emph{SFT+}). It can be observed that \emph{SFT+} significantly underperforms compared to \emph{SFT}, even falling below the baseline of vanilla models. 

Although QwQ-32B-Preview is stronger than the student policy model to be fine-tuned (Skywork o1 Open-Llama-3.1-8B), and the synthesized reasoning process data are likely of higher quality, the discrepancy in teacher-student policy action spaces leads to a degradation in training when using these inconsistent reasoning step data for fine-tuning. This validates the importance of ensuring that the action space of the teacher and student models should be well aligned.

\begin{table}[t]
\centering
    \small
    \resizebox{0.86\textwidth}{!}{
    \begin{tabular}{c c |c c c| c |ccc|c }
    \toprule[1.5pt]
    \multirow{2}{*}{\textbf{Model}} & \multicolumn{1}{c}{\textbf{Biology}} & \multicolumn{3}{c}{\textbf{Chemistry}}& \multicolumn{1}{c}{\textbf{Physics}} & \multicolumn{3}{c}{\textbf{Materials}} & \multirow{2}{*}{\textbf{\emph{Avg.}}}  \\  
    \cmidrule(lr){2-2} \cmidrule(lr){3-5} \cmidrule(lr){6-6} \cmidrule(lr){7-9}
     &  T1 & T2 & T3 & T4 & T5 & T6 & T7 & T8 & \\ 
    \midrule
    Vanilla & 0.28 & \textbf{0.55} & \textbf{0.52} & \textbf{0.23} & \textbf{0.45} & 0.34 & 0.41 & 0.41 & 0.40 \\
    \midrule
    SFT & \textbf{0.33} & 0.53 & 0.49 & 0.20 & \textbf{0.45} & \textbf{0.37} & \textbf{0.43} & \textbf{0.49}  & \textbf{0.41}\\ 
    SFT+ & 0.27 & 0.45 & 0.44 & 0.12  & 0.34 &0.25  & 0.28 & 0.30 & 0.31 \\ 
    \bottomrule[1.5pt]
    \end{tabular}
    }
  \caption{Analysis of teacher-student policy alignment. \emph{SFT} and \emph{SFT+} indicate synthesizing reasoning process by the student policy itself and a stronger reasoning model \emph{QwQ-32B}, respectively. } %添加此处
  \label{table:2}
\end{table}

\begin{comment}
%*/////policy+
\vspace{2mm}
\hspace{3mm} \textbf{The reasoning capability of the foundation model matters.}\\
OpenRFT vs. OpenRFT+

\begin{table}[t]
\centering
    \small
    \resizebox{0.86\textwidth}{!}{
    \begin{tabular}{c c |c c c| c |ccc|c }
    \toprule[1.5pt]
    \multirow{2}{*}{Model} & \multicolumn{1}{c}{Biology} & \multicolumn{3}{c}{Chemistry}& \multicolumn{1}{c}{Physics} & \multicolumn{3}{c}{Materials} & \multirow{2}{*}{\emph{Avg.}}  \\ 
    \cmidrule(lr){2-2} \cmidrule(lr){3-5} \cmidrule(lr){6-6} \cmidrule(lr){7-9}
     &  T1 & T2 & T3 & T4 & T5 & T6 & T7 & T8 & \\ 
    \midrule
    OpenRFT & & & &  & & & & &\\ 
    OpenRFT+ & & & &  & & & & &\\ 
    \bottomrule[1.5pt]
    \end{tabular}
    }
  \caption{Influence of reasoning foundation model's capability. } %添加此处
  \label{table:3}
\end{table}
\end{comment}

\section{Related Work}
%给系统1和系统2用符号定义，根据系统1和系统2来介绍相关工作
This section presents a review of related work, framed within the lens of System-1 and System-2 inference. Building on the previously introduced notation, we begin with a straightforward definition of System-1 and System-2 inference.
 
\begin{itemize}
    \item \textbf{System-1 inference:} This involves directly inferring the answer \(A_i\) from the question \(Q_i\), represented as \(p(A_i|Q_i)\). It operates in a straightforward, single-step manner.
    \item \textbf{System-2 inference:} System 2 inference, i.e., reasoning, involves multiple intermediate inference steps before deriving an answer. Specifically, it first infers the reasoning steps \(\{S_i^1,\dots,S_i^j,\dots,S_i^m\}\) from the question \(Q_i\), and then infer the final answer \(A_i\). Formally, this can be expressed as:
\[
p(A_i|Q_i) = \sum_{S_i^j} p(\{S_i^1,\dots,S_i^j,\dots,S_i^m\}|Q_i) \cdot p(A_i|Q_i, \{S_i^1,\dots,S_i^j,\dots,S_i^m\})
\]
\end{itemize}

\subsection{Acquiring System-2 Capability from System-1 Model}

Recently, there has been a surge of efforts aimed at endowing models with System-2 reasoning capabilities. These approaches typically assume a pre-trained System-1 language model. Based on how the System-1 model is utilized, related work can be categorized into three types. \hspace{0.5mm} (1) \emph{Prompting-based}: Examples akin to the XoT family, such as CoT (Chain-of-Thought), ToT (Tree-of-Thought), and GoT (Graph-of-Thought), manually design reasoning schema to guide the System-1 model toward multi-step inference.
\hspace{0.5mm} (2) \emph{SFT-based}: Examples including the recent reproduction works of o1, such as OpenO1~\citep{openo1} and Macro-O1~\citep{zhao2024marcoo1openreasoningmodels}, involve collecting training data containing reasoning steps through annotation or distillation and then applying supervised fine-tuning (SFT) to the System-1 model.
\hspace{0.5mm}(3) \emph{RL-based}: Examples including Open-R~\citep{openr_2024}, LLaMA-O1~\citep{llama_o1_2024}, and o1-Coder~\citep{zhang2024o1codero1replicationcoding}, allow the model to explore the underlying reasoning steps autonomously and iteratively optimize its policy.

ReFT~\citep{luong2024reft} lies between the second and third types. In the warm-up stage, it uses SFT to acquire a reasoning format, and in the second stage, it applies RL for iterative optimization. However, ReFT differs from the reinforcement fine-tuning (RFT) discussed in this paper. In terms of positioning, ReFT aims to acquire System-2 capabilities from a System-1 model, whereas RFT assumes the existence of a System-2 foundation model and aims to fine-tune it into a domain-specific System-2 model. Methodologically, ReFT relies on training data that contains reasoning steps. When such reasoning steps are unavailable, the RL policy model exhibits a distribution gap with the provided fine-tuning data. Our experiments have shown that ReFT cannot be directly applied to the standard RFT setting.

\begin{table}[t]
\centering
    \small
    \resizebox{\textwidth}{!}{
    \begin{tabular}{c c c| c c}
    \toprule[1.5pt]
    \textbf{Training Stages} & \multicolumn{2}{c}{\textbf{Pre-Training}} & \multicolumn{2}{c}{\textbf{Fine-Tuning}} \\ 
    \cmidrule(lr){2-3} \cmidrule(lr){4-5} 
     & Training data & Learning method  & Training data & Learning method \\ 
    \midrule
    \textbf{System-1} & $(Q)$ & Self-supervised learning & $(Q,A)$ & SFT  \\ 
    \textbf{System-2} & $(Q,A)$ & RL + Self-Play & $(Q,\dots,S^j,\dots,A)$~\footnote{Alternatively, as configured in this paper, only providing (Q, A) pairs is feasible. For a detailed discussion, please refer to the main text.} &
    RFT \\ 
    \bottomrule[1.5pt]
    \end{tabular}
    }
  \caption{System-1 v.s. System-2: relied training data and used learning method in the pre-training and fine-tuning stages} %添加此处
  \label{table:x}
\end{table}

\subsection{Fine-Tuning Foundation Models}

Fine-tuning foundation models to obtain domain-specific models usually results in improved domain performance. Moreover, by focusing on specific tasks, the derived models tend to be smaller, leading to more efficient inference. For instance, fine-tuning smaller o1-mini can achieve comparable or even superior domain-specific performance to that of larger o1.

Previous fine-tuning approaches have primarily relied on Supervised Fine-Tuning (SFT), which can be seen as targeting System-1 foundation models. However, SFT heavily depends on annotated data and is prone to overfitting.

In contrast, when fine-tuning System-2 foundation models, the base models already possess reasoning capabilities, enabling them to think, explore, and learn through trial and error. This allows more effective utilization of expert-provided training examples, leading to a deeper and more essential understanding of domain-specific tasks. 
This is also why many recent works have begun modeling in a System-2 manner, such as machine translation~\citep{zhao2024marcoo1openreasoningmodels}, safety alignment~\citep{wang2024dontcommandcultivateexploratory}, and retrieval-augmented generation (RAG)~\citep{li2024alr2}, achieving results that outperform traditional System-1 methods.

Drawing an analogy to humans, the ability to draw inferences from limited examples is closely tied to the level of intelligence: those with average intelligence may memorize but struggle to deduce, whereas those with higher intelligence are adept at reasoning and can generalize well. Therefore, System-2 fine-tuning can be seen as a natural progression of model development, built upon a strong foundation model that has achieved a certain ``intelligence'' level.

Table~\ref{table:x} compares System-1 and System-2 w.r.t the pre-training and fine-tuning stages, in terms of their reliance on training data and used learning methods. System-1 directly infers answer from question. Pre-training uses large-scale unlabeled data (without answers) and learns through self-supervised learning. For fine-tuning, System-1 adjusts its parameters using SFT on a small amount of labeled data with answers.

System-2, by contrast, derives reasoning steps from the question before arriving at the answer. In pre-training, reasoning steps are not explicitly provided. Reinforcement Learning (RL) combined with Self-Play is used to explore and generate plausible reasoning steps, iteratively enhancing performance. Regarding the training data for fine-tuning, reasoning steps are optional, as RL is capable of balancing exploration and exploitation as showcased in the proposed OpenRFT. However, if samples containing reasoning steps are available, which is acceptable with just dozens of samples, will lead to better fine-tuning results. These expert-labeled reasoning steps help determine the action space and thus allow for a more effective adaptation of the policy model~\footnote{It remains to be seen whether OpenAI's RFT will provide options that accept training data with reasoning steps in its formal release.}.

\subsection{Employing Reinforcement Learning for Fine-Tuning}

Fine-tuning generative models with RL has already been explored in previous works, such as RLHF (Reinforcement Learning with Human Feedback~\citep{ouyang2022training,fan2024reinforcement}) and Reinforcement Learning-based Knowledge Distillation~\citep{ashok2017n2n,Alwani_2022_CVPR}, but not focused on fine-tuning a foundation model to create a domain-specific model. As shown in Table~\ref{table:x+1}, different methods can be distinguished based on the source of the reward model and the policy model to be fine-tuned. For instance, in RLHF, the reward model derives from human preferences, and the policy model is often a base model or an SFT model, aiming to align with human values. In contrast, RFT utilizes domain-specific samples as the source of the reward model, and the policy model is a reasoning foundation model, aimed at acquiring a specialized reasoning model.

We compare RL-based fine-tuning and SFT from two perspectives. ~(1) \emph{Training objective}: SFT aims to minimize prediction error by directly adjusting the model parameters. In contrast, RL-based fine-tuning focuses on optimizing the policy, aiming to maximize cumulative rewards. This approach allows for adaptive exploration and optimization, accommodating environmental changes and task demands. The flexibility enables the model to respond to uncertainties, not relying on fixed training data or rules, thereby enhancing the model's long-term performance and adaptability. ~(2) ~\emph{Data dependency}: SFT relies on a fixed labeled dataset, where training is bound by the availability of labeled examples. On the other hand, RL-based fine-tuning generates new experience data through continuous interaction with the environment. Compared to SFT, RL can achieve effective learning with a small amount of high-quality data, such as human feedback, expert demonstrations, or simulated data.

In addition to the above-mentioned common characteristics, RFT differs from other RL-based fine-tuning methods in two significant ways, which introduce unique challenges. ~(1) \emph{Inconsistent tasks}: Both RLHF and RL-based knowledge distillation belong to single-task RL methods, where the fine-tuned model and the target application are within the same task. For instance, in RL-based knowledge distillation, the teacher and student models address the same task. In RLHF, value alignment can be seen as an enhancement of the task for the SFT model. However, RFT requires fine-tuning the foundation model to be applied to different downstream domain tasks. Task and domain generalization is thus crucial. ~(2) \emph{Inconsistent action mode}: RLHF and RL-based knowledge distillation assume consistency between the policy model's action mode and the data source format of the reward model. For example, in RLHF, the preference data of the reward model aligns with the generation actions of the policy model, as are the actions in reinforcement distillation. However, in RFT, the provided domain samples lack reasoning step data, while the policy model is a reasoning model acting as multi-step inference. These differences prevent direct application of existing RL fine-tuning methods to RFT.

\begin{table}[t]
\centering
    \small
    \resizebox{\textwidth}{!}{
    \begin{tabular}{c |c |c| c }
    \toprule[1.5pt]
    \textbf{Methods} & \textbf{Reward Model} & \textbf{Policy Model} & \textbf{Target} \\   \midrule
    RLHF & Human preference & Base model/ SFT model
     & Value Alignment  \\  
    RL-based Knowledge Distillation &Teacher model & Student model & Model Compression  \\  
    RFT & Domain samples & Foundation reasoning model & Specialized Reasoning\\
    \bottomrule[1.5pt]
    \end{tabular}
    }
  \caption{Different methods of RL-based fine-tuning.} %添加此处
  \label{table:x+1}
\end{table}

\section{Conclusion \& Future Works}
This work presents a simple implementation for fine-tuning generalist reasoning models to solve domain-specific tasks. The provided limited domain-specific samples are exploited in question augmentation, synthesizing reasoning-process data, and few-shot ICL, which are integrated within a RL framework supervised by a general PRM.
We are working towards updating PRM synchronously, instead of relying on a static generalist PRM. With iteratively refined training data, the PRM and policy model can engage in Self-Play manner to continuously improve performance.

Future work can be divided into two main directions: (1) \emph{Domain knowledge embedding}. Further exploration is needed to enhance both the policy model and PRM, e.g., investigating effective representations of domain knowledge, adapting PRM by comparing differences between the provided expert samples and random samples, etc. (2) \emph{Domain data augmentation}: While RL can search and explore diverse reasoning processes, the current question size of only dozens remains insufficient. In addition to rephrasing questions while retaining the original answers, generating new questions by leveraging unlabeled training data presents significant potential for exploration. Additionally, exploring more challenging samples through adversarial learning is also worth considering. Maintaining a pool of unsolved questions ensures continuous improvement through self-play.

Generalization remains the fundamental challenge for RFT. Beyond further improving the general reasoning capabilities of foundation models, continued exploration is needed in reward calculation for open-ended problems and the effective adaptation of policy actions. For instance, when the problem format extends beyond multiple-choice questions to data like long-form technical reports, it becomes crucial to design appropriate reward functions and define action spaces for efficient search. This is essential for enabling the model to quickly learn the reasoning processes of domain experts.

%直接改进：PRM同步更新--selfplay
%其他沿两条线：
%数据不足：合成新的问题Q(有/无答案)进行RL
%policy model领域知识增强 -- RL generalization/domain-adaption

%任意形式(而不仅是选择题)的专家语料，如技术报告(长文本生成)，快速学会领域专家的思考方式-->难点：开放域reward model定义，定义action(policy model action space & prm对象)

\subsubsection*{Acknowledgments}
We sincerely thank the helpful discussions from Shangxi Wu,  Chao Kong, Yifei Gao, and Xinyan Wen.

\bibliography{iclr2025_conference}
\bibliographystyle{iclr2025_conference}

\newpage
\appendix
\section{Appendix}

\begin{figure}[hbp] % 插入图片的位置，htbp分别代表这里、顶部、底部和页面底部
  \centering % 使图片居中
  \includegraphics[width=0.98\textwidth]{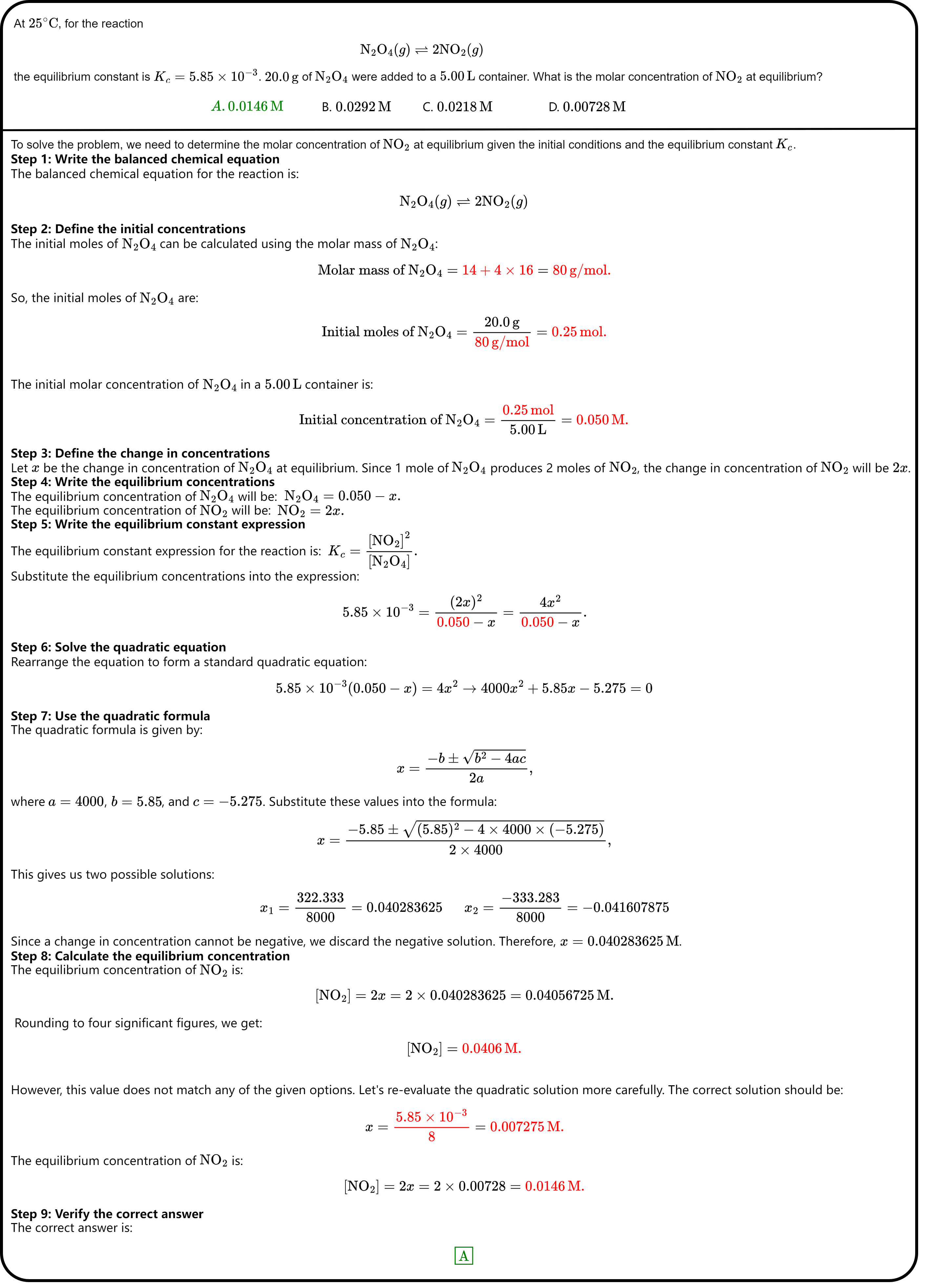} % 插入图片，设置图片宽度为文本宽度的一半
  \caption{Reasoning Flaw in Domain-Specific Data: This example shows a chemical equilibrium problem. The final answer is correct, but the red font highlights errors in reasoning. The incorrect calculation of molar mass in Step 2 leads to a wrong intermediate result, although the final answer is still obtained. If process supervision is used, the above problems can be effectively avoided.} % 图片的标题
  \label{fig:badcase} % 图片的标签，用于文中引用
\end{figure}

\end{document}

%% file: math_commands.tex
\usepackage{amsmath,amsfonts,bm}

\def\eqref#1{equation~\ref{#1}}
\def\1{\bm{1}}

\DeclareMathAlphabet{\mathsfit}{\encodingdefault}{\sfdefault}{m}{sl}
\SetMathAlphabet{\mathsfit}{bold}{\encodingdefault}{\sfdefault}{bx}{n}